\documentclass[10pt, twocolumn]{article}
\usepackage[margin=0.75in]{geometry}
\usepackage{amsmath, amssymb}
\usepackage{graphicx}
\usepackage{listings}
\usepackage{booktabs}
\usepackage{hyperref}
\usepackage{abstract}
\usepackage{algorithm}
\usepackage[backend=biber]{biblatex}
\usepackage{algpseudocode}
\usepackage[toc,page]{appendix}
\newcommand{\lt}{<}
\newcommand{\gt}{>}
\newtheorem{definition}{Definition}

\title{\textbf{Application of curiosity driven exploration methods for hardware interference identification}}

\author{
	Ludovic Matar $^{1}$, Clément Moulin-Frier$^{1}$,Christophe Marabotto$^{2}$,\\ Eric Jenn$^{2}$,Pierre-Yves Oudeyer$^{1}$\\
    \small $^{1}$Flowers AI \& CogSci Lab - National Institute for Research in Digital Science and Technology, Bordeaux, France \\
    \small \texttt{\{ludovic.matar, clement.moulin-frier,pierre-yves.oudeyer\}@inria.fr}\\
    \small $^{2}$IRT Saint Exupery $\cdot$ Institute of Technological Research, Toulouse, France\\
    \small \texttt{\{christophe.marabotto,eric.jenn\}@irt-saintexupery.com}
}

\date{}

\begin{document}
\maketitle

\begin{onecolabstract}
The transition from single-core to multi-core architectures in safety-critical embedded systems introduces significant challenges due to inter-core interference caused by contention for shared hardware resources. Such interference affects execution times and complicates the verification of strict temporal requirements, particularly in domains such as avionics where standards require comprehensive identification of interference sources. Existing interference analysis approaches, whether manual or model-based, struggle to capture the full range of behaviors arising from the complex interactions among micro-architectural components. In this paper, we frame multi-core interference analysis as the exploration of a complex system behavior space. We propose the use of curiosity-driven exploration algorithms from artificial intelligence to systematically and efficiently cover the space of possible interference behaviors. Using a simulator-based environment, we show that the proposed approach achieves broader and more uniform behavioral coverage within a limited experimental budget compared to traditional pseudo-random program generation methods.
\end{onecolabstract}


\section{Introduction}
\label{sec:typesetting-summary}
The shift from single-core to multi-core architectures is essential in safety-critical embedded
systems in multiple domains, such as aerospace and automotive, driven by both the need
to enhance processor performance for increasingly demanding applications and adaptation
to recent technology. However, this transition introduces new complexities due to tasks
running in parallel competing for shared resources; particularly, hardware contention issues
known as inter-core interference. As the occurrence of interference has a consequence on
the execution time, the use of multi-core processors in critical real-time systems represents a
challenge. Indeed, temporal requirements with high levels of confidence have to be met. In
the avionics domain, the AMC 20-193 \cite{esa} standard requires identifying all sources of interference
that can occur when executing tasks in parallel and addressing them.

\begin{figure}[ht!]
\centering
\includegraphics[width=\linewidth]{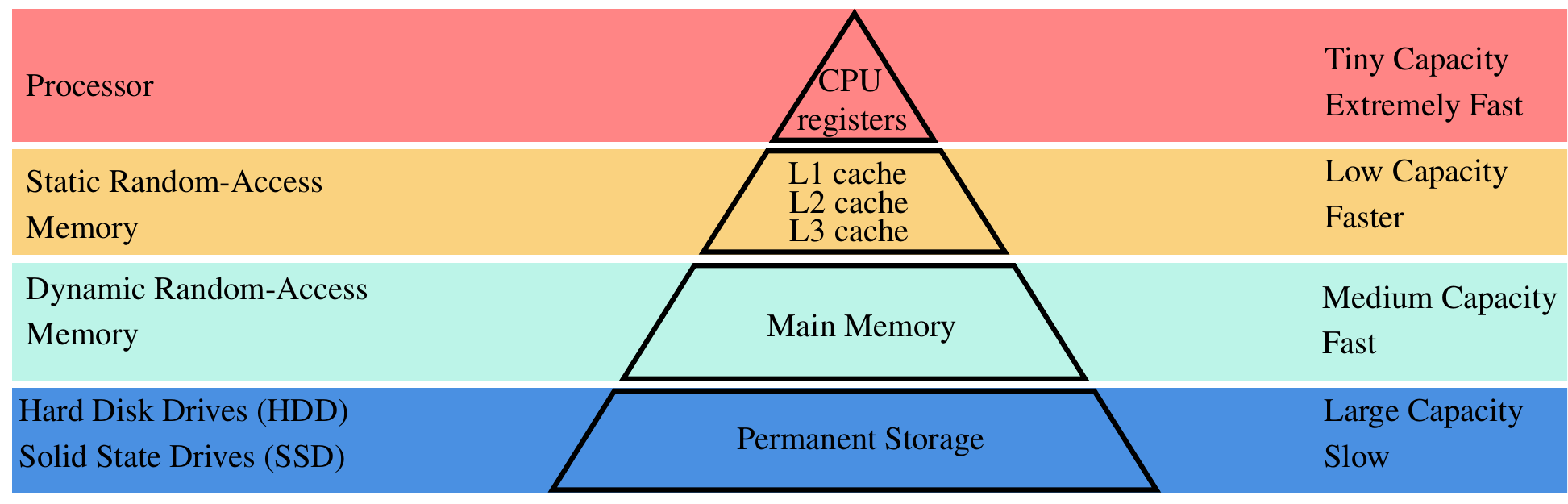}
\caption{Common computer memory hierarchy \cite{Kuldeep}}
\label{hierarchy}
\end{figure}

Systems on chips face a general trade-off between the necessary time to access information stored in a memory resource and their storage capacity. As an element is closer to the CPU, its access time and storage capacity are smaller. This principle is illustrated in Figure \ref{hierarchy}.
Because they are directly on the CPU, registers typically serve information almost instantly. Their storage capacity is very small, and only essential information is stored. When a core treats a memory instruction, it requests the main memory by default. However, the main memory often operates orders of magnitude slower than the CPU. Thus the cache, namely random static memory locates between the registers and the main memory. It has larger latencies than the registers, but remains considerably faster than main memory. Moreover, it has a higher storage capacity than the registers but smaller than the main memory. It stores areas of the main memory that are the most likely to be referenced. 
The main memory consists of DRAM cells that offer a great storage capacity. Therefore, there may be a noticeable slowdown each time the CPU needs to access the main memory.
As a consequence of this organization, competition for accessing stored data and instructions within shared caches and shared memory between CPU cores increases the execution time, and thus creates potential interference channels.
\begin{definition}
Interference is a phenomenon such that, for identical initial conditions, the execution time of an application $S_1$ running in isolation $(S_1,\_)$ on a platform differs from the execution time of $S_1$ running with an application $S_2$ on the platform $(S_1,S_2)$.
\end{definition}
Interference arises from contention on shared resources, e.g., when multiple cores compete to allocate information in cache or memory. Another case is the competition for accessing a shared bus, resulting in bandwidth limitations and delays in data transfer. To summarize, we can distinguish distinct categories of interference occurring because of contention on shared resources \cite{9714355}:
\begin{itemize}
\item \textbf{Cache contention:} When multiple cores compete for space in shared caches, evicting each other’s data and causing cache misses.
\item \textbf{Memory contention:} When multiple cores access a shared memory simultaneously
(such as the main memory), leading to a bottleneck due to limited memory bandwidth.
\item \textbf{Bus contention:} When cores compete for access to a shared bus, resulting in bandwidth limitations and delays in data transfer.
\item \textbf{I/O Contention:} When multiple processes try to perform I/O operations at the same time, leading to delays and reduced I/O performance.
\end{itemize}

As there are multiple types of interference, there exist various types of interference sources to discover. In the following, we will briefly present existing methods that tackle real-time system challenges.

Various methods have been proposed to identify mechanisms that are sources of interference \cite{gonzalez:tel-05025151,boniol:hal-02441323,courtaud:tel-03429679}.  Classic interference analysis methods are either manual and empirical approaches or model-based approaches. The first ones involve manually analyzing the hardware documentation or further characterizing the hardware through experiments and measurements \cite{bin:hal-02271379, girbal:hal-02278292, DBLP:journals/corr/abs-2101-02204, courtaud:tel-03429679, guyomarch:tel-03793814}. The others involve identifying interference from a model of the system through two principles: (1) capturing the hardware-level knowledge about the system to produce a model, and (2) using an analyzer tool which identifies the interference channels from the model \cite{boniol:hal-02441323, bieber:hal-01700857, boniol:hal-03761937}. Classical approaches to interference identification rely heavily on analyzing hardware datasheets, which are often large \cite{2023_hauser_tool-automatically-extracting-hardware}, complex, incomplete, or inconsistent. This makes the process difficult and error-prone, leading to uncertain results and possibly missed interference channels. This can lead, for example, to creating oversimplified models or test cases, resulting in missed interference scenarios or channels \cite{CARLETTI2025103487}.

While existing approaches have enabled better anticipation of potential sources of interference, they remain limited in their ability to cover the entire space of possible behaviors of multi-core processor systems. Interference patterns arise from the complex non-linear interactions between a large number of micro-architectural components. In consequence, they are extremely sensitive to small variations in the concurrent programs that are being executed, which strongly limits the ability to predict the behavior of the systems analytically. Moreover, programs that are generated with pseudo-random procedures are unlikely to cover the entire space of possible behaviors. In this sense, multi-core architectures can be considered complex systems.

A complex system is a system consisting of multiple entities interacting non-linearly with each other, giving rise to emergent phenomena that are difficult to predict \cite{Ladyman2013}. Examples of complex systems in Nature are water molecules crystallizing into snowflakes, collective animal behavior forming complex patterns (e.g., bird flocks), or chains of amino acids forming complex 3D protein structures. To study such systems, scientists often need to build a model mapping their input parameters (e.g., a sequence of amino acids) to their corresponding observed behavior (e.g., the resulting 3D shape of a protein). Complex systems often suffer from so-called \emph{butterfly effects}, where two slightly different initial conditions do not necessarily induce similar outcomes. Such systems also usually exhibit attractor effects, that is, a tendency of the system to evolve towards a particular set of states. Understanding and predicting the behavior of complex systems often requires uncovering the diversity of behaviors they can produce. This is usually a challenging problem, in particular when the parameter space is high-dimensional and when the mapping from the parameter space to the behavior space is highly non-linear. In particular, random exploration of the parameter space is usually inefficient for uncovering the diversity of potential behaviors in most complex systems. 

Recently, contributions in AI have proposed algorithms able to efficiently cover the space of behaviors of any complex system \cite{etcheverry:tel-04504878}, namely, \emph{curiosity-driven exploration algorithms}, which belong to the family of automated discovery algorithms. These algorithms have proven to be very efficient at uncovering a wide diversity of behaviors in many complex systems from different scientific domains, including computer science \cite{DBLP:journals/corr/abs-2007-01195}, physics \cite{PhysRevResearch.6.033052}, chemistry \cite{doi:10.1126/sciadv.aay4237} and biology \cite{10.7554/eLife.92683}. 
In this paper, we propose, for the first time to our knowledge, to apply these algorithms to the problem of discovering interference patterns in multi-core processor systems. For this aim, we use a multi-core processor simulator, enabling us to prototype our methods without having to deal with the complexity of a real hardware architecture. 

We show that our proposed method enables better discovery of the space of possible behaviors of the system with a limited experimental budget, compared to more standard methods based on the pseudo-random generation of programs.

\begin{itemize}
\item In section {section motivation}, we recall general principles regarding interference.
\item Then in section \ref{realtimesystems}, we highlight previous related AI works on real-time systems.
\item In section \ref{explorationstrategy} we unfold our exploration strategy step by step.
\item In section \ref{results}, we present a summary of our findings, with results of diversity measures.
\end{itemize}

\section{Related works}
\label{realtimesystems}
The interest in AI techniques within the real-time computing community is relatively recent. Although studies tackling interference on embedded platforms using AI-based techniques exist, to our knowledge, no studies have explicitly targeted their use to identify sources of interference.
Nevertheless, we will review the key works in these related areas and explore how they might be adapted or inspire solutions for our problem.

Several works have focused on the use of Machine Learning to estimate Worst-Case Execution Times, whether for single-core platforms \cite{amalou:tel-04406029} or multi-core ones \cite{bonenfant:hal-03116285}. An interesting work closer to our objective is the Kryptonite approach \cite{Singh23}, introducing a framework that synthesizes a maximally interfering environment for a Program Under Test (PUT) executing on a multi-core platform. In other words, the Worst-Case Program Interference is targeted. The approach relies on a set of code snippets, called gadgets, to hammer specific shared resources in the multi-core platform and create interference. Kryptonite then arranges the gadgets in order to maximize interference in two phases: firstly, a greedy approach is used to iteratively build a sequence of gadgets that increases execution times of the PUT. And secondly, a Reinforcement Learning (RL) algorithm is used to fine-tune the interfering environment. Although the objective is not the same, this approach motivates the idea of using an automated discovery algorithm on a multi-core platform to explore interference as it requires many interactions with the system. Indeed, a complex system is exploited, and the objective of reaching an optimum in an output space by exploring a code space is fulfilled. In contrast with an optimization method that minimizes a loss function, we propose a space coverage technique to induce a large diversity of behaviors in a given complex system.
In the next section, we present our basic simulator upon which we will perform our exploration.

\section{Description of our dual-core model}
We justify the implementation of the simulator and briefly describe components and their behaviors.
A minimum of two cores with shared memory components is, by definition, necessary to induce interference phenomena. Thus, our basic simulator model contains:
\begin{itemize}
\item two cores,
\item their respective private L1 caches,
\item a shared L2 cache unit and a DDR memory with its controller,
\item an interconnect model that transmits instructions sent by the cache to the memory.
\end{itemize}
\begin{figure}[ht!]
\centering
\includegraphics[height=5cm]{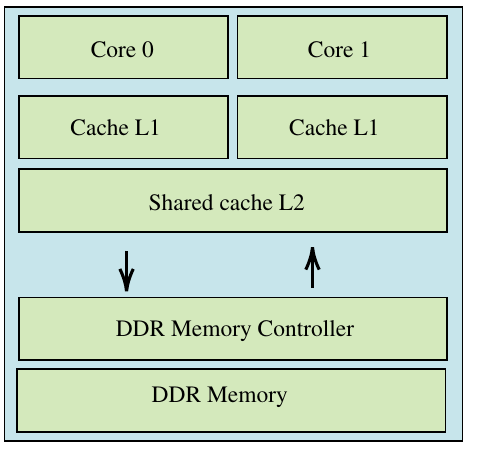}
\caption{Simulated dual-core architecture used in this paper. Each core has a private L1 cache and shares an L2 cache with the other. The L2 cache is connected to the main DDR memory via its DDR memory controller.}
\label{simulatorillustration}
\end{figure}
With such a configuration, interference arises in shared components, such as the L2 cache, the DDR memory controller, and the interconnect. In the following, we explain the functioning of some of the basic blocks of our simulator.

\subsection{Core model}
\label{coremodel}
As interference is mostly due to memory mechanisms, the simulation of cores handles exclusively memory access instructions; that is, the instruction set is reduced to read and write operations in memory since, ultimately, only memory accesses are simulated. In the following, we denote the "read" operations by RD and the "write" operations by WR.

In our algorithms, for each core, a simple loop models the "fetch and execute" cycle. At each CPU cycle, the core can:
\begin{enumerate}
\item wait for the end of an RD operation,
\item execute an RD, 
\item execute a WR, 
\item execute an instruction that does not perform memory accesses, which, in the case of the simulator, amounts to doing nothing.
\end{enumerate}

As in real-life applications, we make use of a set of virtual addresses that maps to the set of physical addresses, that is, elements of the micro-architecture memory. Within our programs, virtual addresses are simply modeled as numbers. The function that associates a virtual address to a location in the DDR memory is given in \ref{appendix:graphddr}.

Within the pairs, the programs do not exchange data to be written or read, but are to be executable and interpretable; moreover, this data is not modeled as it does not play any role in interference phenomena, as opposed to instructions. Programs consist of RD and WR at given addresses. The shipment is done via the private L1 caches, as instructions are not stored in either the cache or the memory. 
Thus, programs are concretized as simple dictionaries with the format $\{$cycle:(type,address)$\}$ as follows. For each instruction in a program, its key is the cycle at which the considered instruction is sent to the simplified simulator.
\noindent\begin{minipage}{\linewidth}
	\begin{lstlisting}[caption={Example of simplified assembly programs.},basicstyle=\ttfamily\scriptsize]
program_0 = {23:('read', 11),37:('read',17),49:('write',6)}
program_1 = {12:('write', 24),18:('read',39),32:('write',37)}
\end{lstlisting}
\end{minipage}\hfill

The maximum length of a program and the largest cycle for which an instruction is sent are fixed as parameters. 
This model is obviously very simplified compared to the actual operation of a physical target device. Waiting for the completion of a read memory access to continue code execution does not correspond to reality because the processor has various mechanisms specifically allowing it to hide these latencies. This wait is only necessary to preserve true data access dependencies (e.g., a program WR @x RD @x must be executed in this order and be preserved at run time to maintain program semantics).

\subsection{Cache model}
Each cache level is configurable in total size, cache line size, and associativity.
By default, the behavior is of the "write-back" type, that is, writing to lower-level memory occurs when a cache line is evicted. During the write operation, the line is simply marked as "dirty" and is written to the lower-level memory during its eviction.
It is also possible to employ a "write-through" behavior in which writing to the lower-level memory is done immediately.
The management of the eviction of cache lines is carried out by a PLRU whose role is to determine the line to be replaced based on the current state of the cache. Ideally, we would like to implement a behavior of the LRU (Least Recently Used) type, which would consist of eliminating the line used least recently in order to make the most of the locality principle \footnote{According to the spatial locality principle, data is very likely to be accessed when one of its neighbors has recently been accessed, whereas according to the temporal locality principle, recently accessed data is very likely to be accessed again.\cite{courtaud:tel-03429679}}. However, this strategy is expensive to implement and we often prefer to use a simpler mechanism called Pseudo-LRU, which relies on a binary tree.

Thus, this algorithm includes a function that maintains the data structure with a binary tree and which allows choosing the next row to be evicted based on memory accesses. It also includes a function allowing the choice of the next cache row to be evicted using the information contained in the binary tree. The last level cache (L2 in figure \ref{simulatorillustration}) is shared by both memory hierarchies.
There is no cache coherency management mechanism.
\subsubsection{DDR model}
\begin{figure}[ht!]
\centering
\includegraphics[width=7cm]{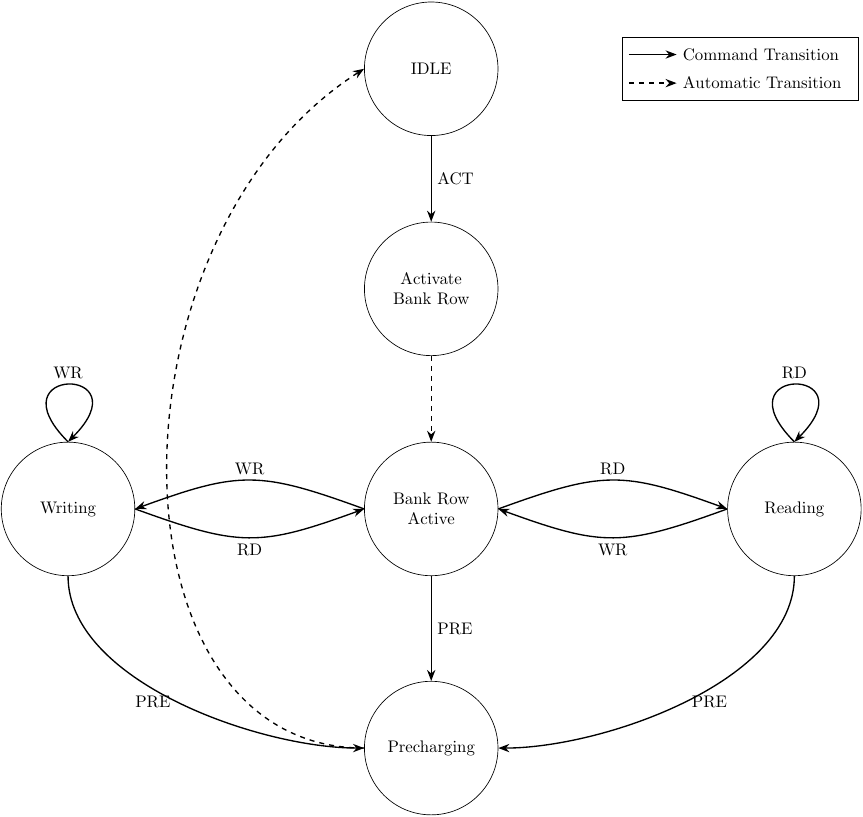}
\caption{Simplified state diagram assuming a single bank \cite{gonzalez:tel-05025151}}
\label{statemachine}
\end{figure}

Let's recall that DRAM is organized in rows and columns of banks. Each bank contains an additional row called a row buffer. When information is accessed, its row is stored in the row buffer.

In our model, the memory consists of several banks, each of which contains a row buffer acting as a cache. 
Latencies are modeled and play an important role in the interference delay in the following way:
\begin{itemize}
\item Access to different rows from the same bank induces significant delays because of the temporal cost of changing rows in the buffer.
\item Simultaneous accesses to distinct rows from the same bank induce larger temporal costs than simultaneous accesses to distinct banks. In other words, intra-bank interference induces larger delays than inter-bank interference.
\end{itemize}

Each bank is managed by a timed state machine that reproduces the one described in \cite{gonzalez:tel-05025151} (see figure \ref{statemachine}).
The state of the banks of the DDR determines the completion dates of operations. Requests are issued by the DDR controller. Let's note PRE as the operation that deactivates an open row of a given bank; this action writes the selected information from the row buffer to the memory. And ACT is the operation that activates a closed row by storing the information from the given row in the buffer. Additional information on the DDR is given in \ref{appendix:graphddr}.
\subsection{DDR controller model}
The memory controller is in charge of reordering memory access requests to maximize the memory access rate. It implements several queues: a queue for commands, a queue for read data, and a queue for the data being written.
Prioritization rules are modeled. Indeed, the RD are prioritized over the WR in order to minimize waiting times, while preserving memory consistency (WR @x RD @x must be executed in this order).
Instructions targeting lines that are already in row buffers are prioritized in order to avoid the penalty related to completing a PRE=$\gt$ACT sequence. RD and WR are processed in "batches" to avoid the overhead related to the transition between RD and WR. Mechanisms of our DDR controller are described in \ref{appendix:ddrcontroller}.

\section{Overview of the exploration strategy}
\label{explorationstrategy}
\begin{figure}[ht!]
\centering
\includegraphics[width=1.0\linewidth]{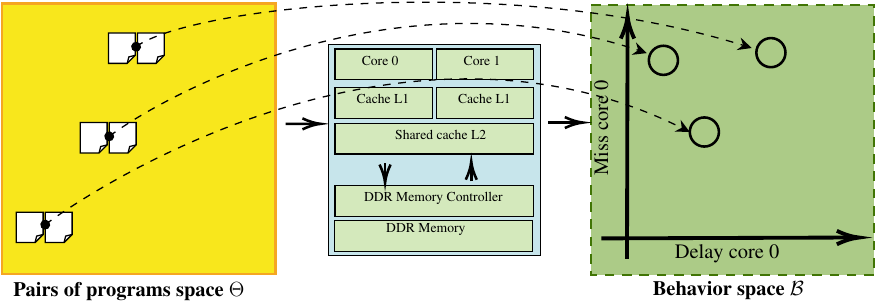}
\caption{Illustration of a complex system associated with multi-core hardware. On the left side of the figure, pairs of simplified assembly programs are represented in the parameter space $\Theta$. Each of them is executed on the platform and corresponds to an outcome (behavior) on the right-hand side, representing the behavior space $\mathcal{B}$. In our case, both the parameter space and the behavior space can be high-dimensional spaces, although in general, the parameter space has a larger dimension than the behavior space.}
\label{simulatorillustration}
\end{figure}

The Intrinsically Motivated Goal Exploration Process is an instance of automated discovery algorithms. This diversity-driven strategy aims to maximally cover a behavior space. During its exploration, a discovery agent iteratively samples goals uniformly in the behavior space and offers appropriate parameters to reach them by leveraging previously found (parameters, outcome) relations. The diversity maximization is a side effect of the goal exploration process. The exploration consists of several episodes called "experiments," during which the agent continually improves its ability to discover new singularities. 

For interference identification, the considered complex system is the simulated hardware platform. In contrast to Kryptonite, for example, which attempts to minimize/maximize interference delays, an IMGEP instead aims to discover a large diversity of possible interference patterns. We wish to explore a behavior space $\mathcal{B}$ that characterizes interference patterns within the hardware by manipulating inputs from a parameter space $\Theta$; see Figure \ref{simulatorillustration}. In our case, we establish a parameter space made of simplified assembly programs. It is essential, on the one hand, to formally model basic elements of the exploration, such as the parameter space $\Theta$ and the behavior space $\mathcal{B}$, and on the other hand, to specify the agent's internal models.

\begin{definition}
We call a "pair of programs" a couple of two programs of code $S_1$ and $S_2$ allocated respectively to the $C_1$ and $C_2$ cores of an execution platform. $S_1$ and $S_2$ are independent code programs. In particular, they do not exchange data.
\end{definition}
\begin{definition}
We call an "experiment" the interaction with our complex system. It involves the parallel execution of the programs $S_1$ and $S_2$, and the execution of $S_1$ and $S_2$ alone.
All experiments are to be independent; this involves, for example, clearing the cache, DDR, and interconnect before every execution. 
\end{definition}

\begin{figure*}
\centering
\includegraphics[width=.7\textwidth]{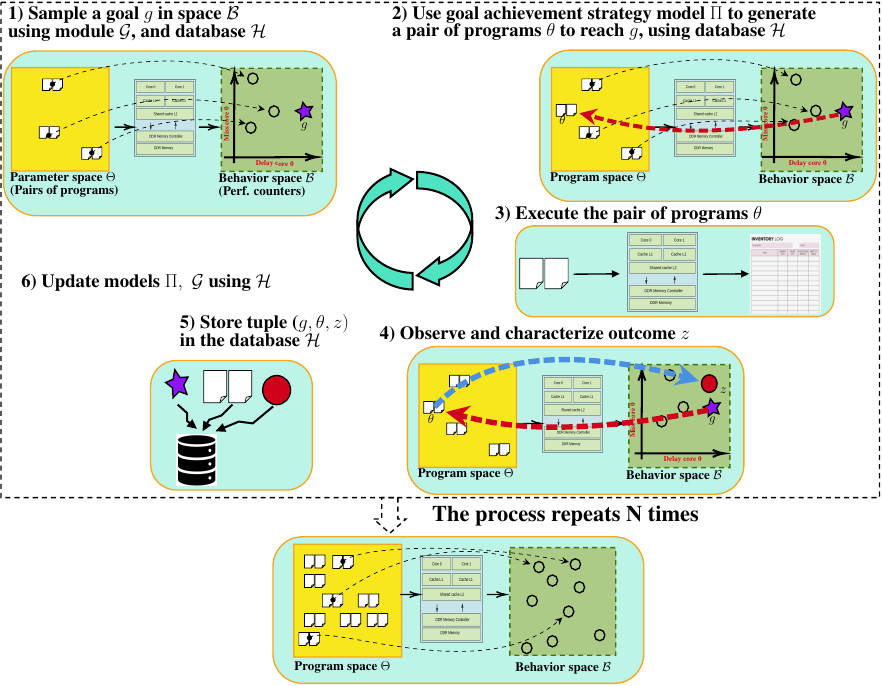}
\caption{Schematic view of an IMGEP for our case of interference identification, based on the general description of IMGEP $\ref{appendix:IMGEP}$.}
\label{defexplorscheme}
\end{figure*}
From an IMGEP perspective, an element of a behavior space is seen as a goal to achieve; thus, goals are target behaviors in the behavior space. In our experiments, a goal is a vector encoding a targeted interference pattern. The discovery agent samples goals to solve and uses its strategy to solve them. Two internal models are used for these respective aspects: one for the goal generation strategy $\mathcal{G}$, and a goal strategy achievement model $\Pi$. As illustrated in Figure \ref{defexplorscheme}, the agent fills a database $\mathcal{H}$ to update its internal models during the exploration, and thus the acquired data is reused to extract potential solutions to solve other goals. 

An IMGEP is a loop algorithm that iterates online by interacting with the considered complex system. The algorithm starts with a warming set of $N_{\mbox{init}}$ iterations. This helps the exploration to escape poor knowledge regions.

This type of architecture highlights previous independent experiments stored in the database $\mathcal{H}$. Using efficient sampling methods, this algorithm allows one to quickly reach a great diversity of outcomes in the behavior space. Moreover, its usage of the database prevents forgetting during exploration.

\subsection{Parameter space}

A pair of programs is required to run each experiment. As presented in Figure \ref{defexplorscheme}, the discovery agent generates a pair $\theta\in\Theta$ at each iteration. Following the general program constraints discussed in part \ref{coremodel}, we recall that most sources of interference are due to memory mechanisms; therefore, we choose to produce pairs of simplified assembly instructions to constitute a parameter space $\Theta$. Indeed, as presented before, only memory instructions RD and WR are modeled.
To get closer to a real-life application, we can model two distinct sets of addresses for the cores, i.e., core 0: 0,20 and core 1: 21,40. As two distinct addresses can map to the same physical address, it will still be possible to observe interference.

With the maximum program length and largest shipment cycle being carefully chosen, we synthesize a parameter space of a reasonable cardinality to thus have a better understanding of the results.
\subsection{Behavior space}
\label{behaviorspace}
All microarchitectural mechanisms are known, as the simulator is a white box. We wish to identify the ones responsible for interference. A set of relevant performance counters provides building blocks for the behavior space; see table \ref{perfsimulator}.
\begin{table}[ht!]
\centering
\resizebox{1\linewidth}{!}{
\begin{tabular}{|l|l|l|l|}
\hline
\textbf{Performance counter} & \textbf{Category} & \textbf{Description}             \\ \hline
Processor cycles             & General           & Number of executed cycles            \\ \hline
Instructions completed        & General           & Number of completed instructions     \\ \hline
Decode stalled               & General           & Number of cycles in a waiting status \\ \hline
Cache misses                 & Cache             & Number of cache misses L1/L2           \\ \hline
Cache store allocates        & Cache             & Number of line allocations in L1/L2  \\ \hline
Cache demand access          & Cache             & Number of requests for L1/L2         \\ \hline
DDR store misses & DDR & Number of DDR stores \\ \hline
DDR load misses        	     & DDR               & Number of misses		            \\ \hline
DDR demand access            & DDR               & Number of requests                   \\ \hline
DDR store allocates          & DDR               & Number of line allocations           \\ \hline
\end{tabular}
}
\caption{Set of available performance counters in the dual-core simulator}
\label{perfsimulator}
\end{table}

Such performance counters can be clock cycles, row misses, instruction types, branch mispredictions, and the number of stalls. Indeed, the phenomena quantified by those performance counters allow for the description of conflicts occurring in shared resources and thus are directly related to interference. In general, information on the platform will always be needed in a context where interference is being addressed. If a platform is a black box, all the requirements of the certification file cannot be fulfilled.

Each element of the established behavior space is perceived as a potential goal for the discovery agent. A goal is defined as a pair $g = (z_g, \mathcal{L}_g)$ with an embedding $z_g\in\mathcal{B}$ and an evaluation function $\mathcal{L}_g = \mathcal{D}(\cdot,z_g)$.
Ideally, $\mathcal{B}$ has "good properties" such as being a low-dimensional metric space. The metric property is important for the goal achievement strategy of the discovery agent. Indeed, one wants to evaluate the distances between some previous experiment outcomes $z$ and a goal embedding $z_g$. For example:  $\mathcal{L}_{g}(z) = {||z - z_g||}_2$. This aspect is essential for designing an efficient goal achievement strategy, as one wants to efficiently select previous experiment outcomes and their associated parameters that are potentially suited for the generation of new candidate parameters.

\begin{table}[ht!]
\centering
\resizebox{1.0\linewidth}{!}{
\begin{tabular}{|l|l|l|l|}
\hline
\textbf{Features} & \textbf{Category} & \textbf{Description}             \\ \hline
Processor cycles diff core  & General     & Difference $[\mbox{nb of executed cycles}][\mbox{non-iso}] - [\mbox{nb of executed cycles}][\mbox{iso.}]$\\ \hline
Processor mutual & General     & Difference $[\mbox{nb of executed cycles}][\mbox{non-iso. core 0}] - [\mbox{nb of executed cycles}][\mbox{non-iso core 1}]$ \\ \hline

Cache write misses           & L2 Cache    & Difference $[\mbox{nb of L2 cache misses}][\mbox{non-iso}] - [\mbox{nb of L2 cache misses}][\mbox{iso.}]$\\ \hline
Cache write hits             & L2 Cache    & Difference $[\mbox{nb of L2 cache hits}][\mbox{non-iso}] - [\mbox{nb of L2 cache hits}][\mbox{iso.}]$\\ \hline
Cache read misses           & L2 Cache    & Difference $[\mbox{nb of L2 cache misses}][\mbox{non-iso}] - [\mbox{nb of L2 cache misses}][\mbox{iso.}]$\\ \hline
Cache read hits             & L2 Cache    & Difference $[\mbox{nb of L2 cache hits}][\mbox{non-iso}] - [\mbox{nb of L2 cache hits}][\mbox{iso.}]$\\ \hline
DDR misses             & DDR         & Difference $[\mbox{nb of misses}][\mbox{non-iso}] - [\mbox{nb of misses}][\mbox{iso.}]$ for each pair (bank, row)\\ \hline
DDR hits        	     & DDR         & Difference $[\mbox{nb of hits}][\mbox{non-iso}] - [\mbox{nb of hits}][\mbox{iso.}]$ for each pair (bank, row)\\ \hline
\end{tabular}
}
\caption{Features constituting the behavior space $\mathcal{B}$. For any performance counter $C$, we calculate two features $F_{C}$ as non-iso. vs. iso. differences $F_{C} \equiv C[\mbox{non-iso}] - C[\mbox{iso.}]$. "iso." corresponds to an execution with only one of the two cores, 0 or 1.}
\label{counters}
\end{table}
Because interference arises following micro-architectural behavior conflicts, we target shared resource performance counter differences in values between the executions in isolation (\textbf{iso}) and the executions in non-isolation (\textbf{non-iso}). Given a pair of programs, we calculate for each selected performance counter its difference when executing in isolation on one core and in non-isolation on both cores. The features in table \ref{counters} constitute the behavior space $\mathcal{B}$.

With $n_{\mbox{banks}}$ banks and $n_{\mbox{rows}}$ rows, the dimension of $\mathcal{B}$ is given by $n = 3+4 \cdot 2 + 2\cdot 2\cdot n_{\mbox{banks}}\cdot n_{\mbox{rows}}$ and $\mathcal{B}\subset\mathbb{R}^{n}$. If $n_{\mbox{rows}}=3$ and $n_{\mbox{banks}}= 4$, we then have $\mathcal{B}\subset\mathbb{R}^{59}$. The performance of the exploration algorithm might decrease in case of a large behavior space dimension as a consequence of the curse of dimensionality, where the interpretation of distances loses its sense. Therefore, working with a parsimonious set of data might imply better results in terms of diversity measures. This chosen configuration of a behavior space is an Euclidean space. Therefore, we can measure distances between the current goal embedding $z_g$ and known features $z$ in our database.
Let us remark that the performance counters associated with the features presented in table \ref{counters} are not systematically available when working with a physical device.

\subsection{Goal sampling}

\begin{figure}[ht!]
\centering
\includegraphics[height=3.0cm]{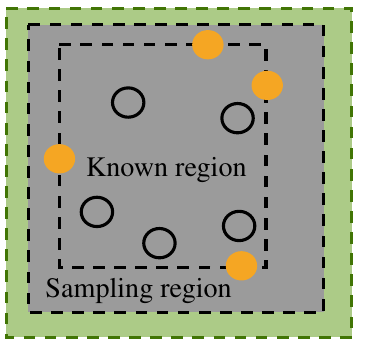}
\caption{Schematic view of a sampling process. Yellow points represent border outcomes from the already known region and allow the determination of the sampling region for the next experiments.}
\label{sampling}
\end{figure}

For the discovery agent, the concept of "curiosity" is materialized by the sampling of goals. For the sampling of goals, the simplest case scenario is to sample them uniformly at random in a connex subregion of $\mathcal{B}$ including the set of already discovered points. As illustrated in Figure \ref{sampling}, choosing the sampling region slightly larger than a subregion including the set of already discovered points allows to target regions of $\mathcal{G}$ that have not been reached yet but are close to the already discovered ones.

We periodically set the sampling boundaries based on the database $\mathcal{H}$, and we simply sample goals with a uniform distribution. For instance, we determine the minima and maxima of the discovered values for each 1-dimensional feature:
\begin{itemize}
\item  ${min}_{\mathcal{B}} g:= (min  \{z_1,z\in\mathcal{H}\},\cdots,min \{z_{\mbox{dim}(\mathcal{B})},z\in\mathcal{H}\})$
\item  ${max}_{\mathcal{B}} g:= (max  \{z_1,z\in\mathcal{H}\},\cdots,max \{z_{\mbox{dim}(\mathcal{B})},z\in\mathcal{H}\})$
\end{itemize}
Then, to establish the slightly larger sampling region, we fix two factors, e.g., $f_1 = 0.8, f_2 = 1.2$, that determine percentages of the previous minimal and maximal values for the new uniform distribution that will be used to sample goals $g\in\mathcal{B}$. Since $\mathcal{G}$ is a multidimensional space, the resulting distribution will be the cartesian product of several 1-dimensional uniform distributions:
$$z_g\sim\mathcal{U}([f_1 \cdot {({min}_{\mathcal{B}}g)}_1 ,f_2\cdot {({min}_{\mathcal{B}}g)}_2])\otimes\cdots$$
$$\cdots\otimes\mathcal{U}([f_1\cdot {({min}_{\mathcal{B}}g)}_{\mbox{dim}(\mathcal{B})},f_2 \cdot {({max}_{\mathcal{B}}g)}_{\mbox{dim}(\mathcal{B})}])$$
\subsection{Goal achievement strategy}

\begin{figure}[ht!]
	\centering
	\includegraphics[width=1.0\linewidth]{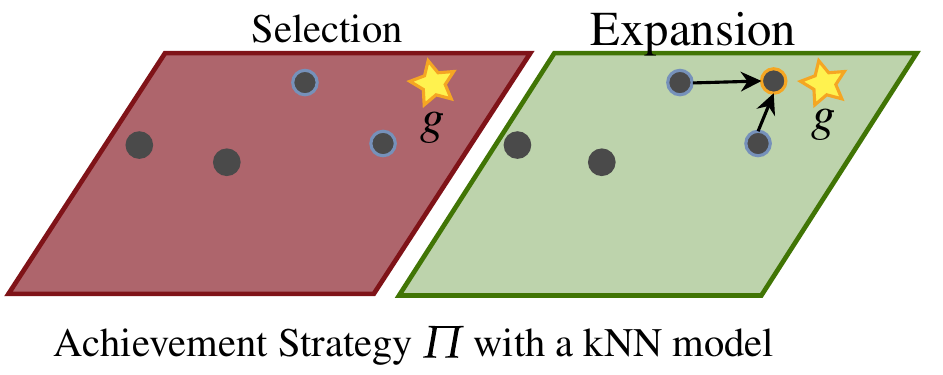}
	\caption{Illustration scheme of a goal achievement strategy. The left hand side illustrates the selection of outcomes stored in the database that will allow expansion, that is the synthesis of a new outcome closer from $g$ (right hand side). Here, the two closest neighbors to $g$ in the database $\mathcal{H}$ are selected, and their parameters are used to obtain the new outcome closer to $g$.}
	\label{strategy}
\end{figure}

We present our population-based achievement strategy, which resembles an evolutionary algorithm, where the set of pairs of programs together with their respective outcomes is seen as a population. Thus, our algorithm implies the performance of mutations of the individuals.

At each step during an IMGEP exploration, the programs (parameters) and outcomes $(({S_1},{S_2}),z)\in\Theta \times\mathcal{B}$ that make up the agent's discoveries are stored in the database $\mathcal{H}$.
Because the \textbf{exploration budget} is low, we choose the agent to perform non-parametric learning, which means that the complexity of the model grows linearly with the size of the database $\mathcal{H}$. 

The goal achievement strategy is typically made of two steps with two respective operators. During the \textbf{selection} step, the discovery agent selects outcomes. The \textbf{expansion} step consists of using their corresponding parameters to produce another one, leading to an outcome close to the goal $g$, see figure \ref{strategy}.

We use a $kNN$ \footnote{Let's recall that $kNN$ is a supervised machine learning model that predicts a label $y$ of a given vector $x$, using the $k$ closest neighbors of $x$ associated labels.} model as a selection operator. This selection operator relies on an expansion operator, as kNN \textbf{selects} $k$ promising outcomes from $\mathcal{H}$  and \textbf{expands} them by producing a candidate pair of programs to achieve $g$. In particular, we select the $k$ closest outcomes to the given goal $g$ within the database $\mathcal{H}$ using a well-designed evaluation function $\mathcal{L}$.  We then mix the $k$ corresponding pairs of programs together to obtain a candidate pair of programs to achieve goal $g$. 

We establish a weighted evaluation function so that all features are equally considered in the selection step:$$\mathcal{L}_{g}(z) = \sum_{F}\frac{1}{\mbox{max}_{F} - \mbox{min}_{F}}\cdot {(z_{F}-g)}^{2},\forall z,g\in\mathcal{B}$$

For mixing the programs, we select contiguous parts from multiple programs, preserving timing, and ensuring the resulting program fits within the fixed maximum cycle. The length of each segment is chosen at random. The key parameters are the \textbf{number of segments to produce} the output program and the \textbf{number of programs to mix}.

To ensure more efficiency, one  also apply a \textbf{mutation operator} following the mixing operator. The mutations performed are $\colon 1.$ randomly change existing instructions, $2.$ delete random instructions, $3.$ add random instructions. 
Consequently, different patterns of the programs change, such as the number of instructions, number of memory accesses, and the set of used addresses.

In the particular case of $k=1$, we simply select the closest observation $z$ of $g$ in the database $\mathcal{H}$ and apply the \textbf{mutation operator} to obtain the candidate parameter. The mixing and mutation operators are detailed in \ref{appendix:mixing} and \ref{appendix:mutation}.

During the first iterations of the algorithm, IMGEP typically outputs outcomes that are not diverse. To overcome the difficulty of the agent exploring during its first iterations, a warm-up made of several random iterations is performed. The agent initially induces poor diversity, and while the population stored in the database $\mathcal{H}$ grows, the agent is more likely to solve its goals. Hence, on average over time, goals targeted by the agent are not reached with the parameters $\theta$ outputted by IMGEP; meanwhile, it is a powerful method that leads to high-diversity discoveries.

\section{Results}
\label{results}
We run IMGEP for a total of $N=10^{4}$ iterations, and we initialize it with a warming set of $N_{init} = 10^{3}$ iterations of random program synthesis. We choose to run it with $k=1,2$ and $3$ as parameters for the $kNN$ model of the goal achievement strategy $\Pi$. Each program contains between $1$ and $10$ instructions. The maximum cycle for the last instruction to be sent is $60$. As said previously, we model two distinct sets of addresses for the cores, i.e., core 0: 0,20 and core 1: 21,40. The selection of the simulator parameters chosen to run our experiments is available in \ref{appendix:graphddr}.

We first assess the diversity of the resulting datasets through diversity measures. Secondly, we explore the results and attempt to infer diagnostics from the data. Finally, we discuss approaches that would be more relevant for IMGEP.
\subsection{Diversity evaluation}
\begin{figure}[ht!]
	\centering
	\begin{minipage}{\linewidth}
		\centering
		\includegraphics[width=\linewidth]{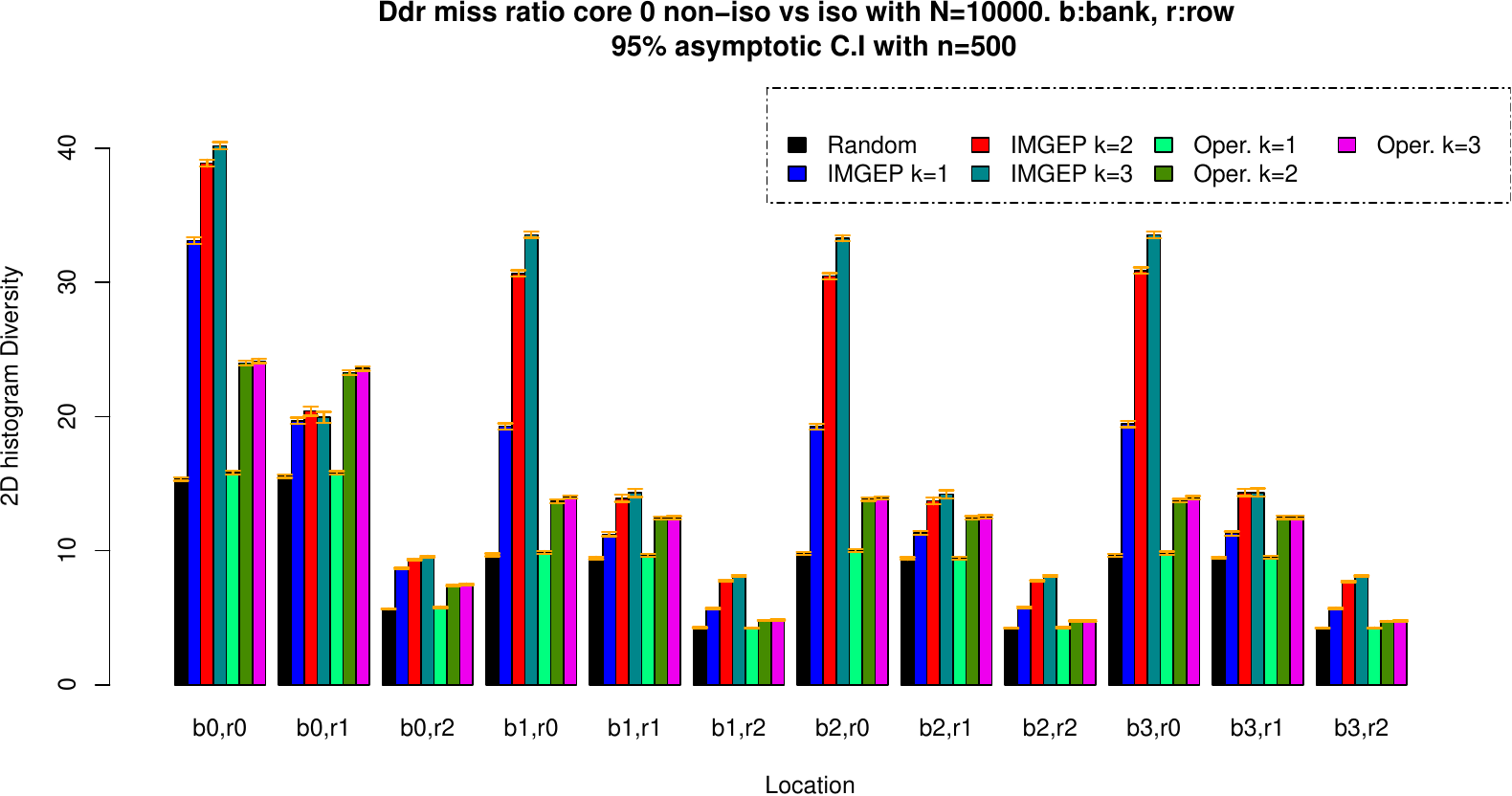}
	\end{minipage}
	\begin{minipage}{\linewidth}
		\centering
		\includegraphics[width=\linewidth]{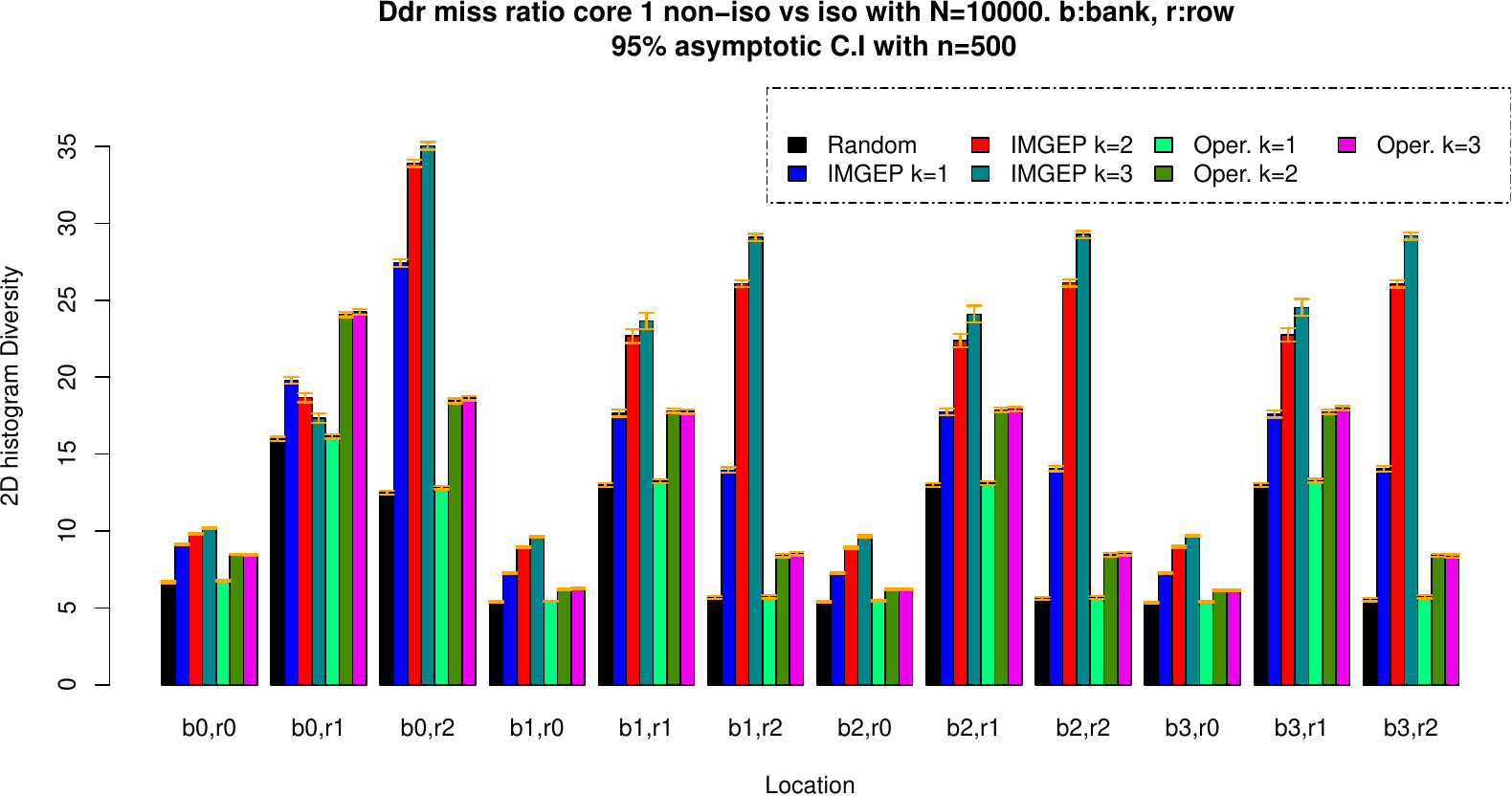}
	\end{minipage}
	\caption{Diversity of miss ratio for both cores and all bank row pairs. Orange bars represent lower and upper confidence bounds using the Student's t-distribution asymptotic confidence interval $\mathbb{P}(\bar{X}- u_{1-\frac{1}{\alpha}}\sqrt{\frac{\hat{v}}{n}} \leq g(\theta)\leq \bar{X} + u_{1-\frac{1}{\alpha}}\sqrt{\frac{\hat{v}}{n}})\longrightarrow .95$ with mean and unbiased standard deviation estimators.}
	\label{missratiosddr}
\end{figure}
We compare our method with two simple strategies: 1. A random exploration, that is, generating pairs of programs randomly. And 2. to be compared with IMGEP for $k$, an algorithm that selects $k$ pairs of programs randomly from its database $\mathcal{H}$, mixes them to obtain a single pair, and mutates it. Indeed, we would like to ensure that a larger IMGEP diversity does not occur simply because of the design of the mixing and mutation operators. For this matter, we aim to demonstrate that mixing and mutating random selections of programs from the same warming set does not lead to a significantly higher diversity than that of IMGEP runs. The mixing and mutation operators are the same as IMGEP's. Hence, for a fair comparison, the first $N_{init}$ iterations of the random exploration are shared by the IMGEP runs and by the operator combination exploration.
We quantitatively assess each exploration process by simply establishing diversity measurements of the outcomes stored in the respective database $\mathcal{H}$. In particular,  we  observe how the behavior space $\mathcal{B}$ is covered by the corresponding outcomes $o_1, \cdots, o_n$.

In the following, we denote two particular subspaces of the behavior space as the \textbf{time behavior space} and \textbf{hit/miss space}. We construct the following measure suitable to assess diversity: The number of bins filled in a $n$-dimensional histogram. With $n = {3+4 \cdot 2 + 2\cdot 2\cdot n_{\mbox{banks}}\cdot n_{\mbox{rows}}}$, being the dimension of each outcome we obtain after an experiment, see part \ref{behaviorspace}. We denote the quantity corresponding to each axis $j$ of an outcome $F$ by $F_j\in\mathbb{R}, 1\leq j\leq n$. The values of $\mbox{inf}(F_{j})$ and $\mbox{sup}(F_{j})$ are carefully chosen so that $\forall F, F\in]\mbox{inf}( F_{j}),\mbox{sup}(F_j)[$, and that for every axis $j$, these values determin the smallest envelopp of possible values.

\begin{table}[ht!]
\centering
\resizebox{\linewidth}{!}{
\begin{tabular}{|l|l|l|l|}
\hline
	\textbf{Performance counter}  & \textbf{Inf} & \textbf{Sup}             \\ \hline
Processor cycles              & 0            	     & Handmade determined upperbound            \\ \hline
Instructions completed        & 0 		     & Maximal number of instructions     \\ \hline
Decode stalled                & 0 		     & Maximal number of instructions\\ \hline
Cache misses                  & 0 		     & Maximal number of instructions\\ \hline
Cache store allocates         & 0 		     & Maximal number of instructions\\ \hline
Cache demand access           & 0 		     & Maximal number of instructions\\ \hline
DDR store misses              & 0 		     & Maximal number of instructions\\ \hline
DDR load misses        	      & 0 		     & Maximal number of instructions\\ \hline
DDR demand access             & 0 		     & Maximal number of instructions\\ \hline
DDR store allocates           & 0 		     & Maximal number of instructions\\ \hline
\end{tabular}
}
\caption{Set of available performance counters in the dual-core simulator}
\label{perfsimulator}
\end{table}

Let us denote a chosen partition of $]\mbox{inf}( F_{j}),\mbox{sup}(F_j)[$ by $x(j), 1\leq j\leq n$ with $n_j$ points, such that $\colon\mbox{inf}{(F_j)} = x(j)_1 \lt\cdots\lt x(j)_{n_j}=\mbox{sup}(F_j)$. 
For a family of partitions $(x(1),\cdots,x(n))$, we define a bin as :
$$bin((x(1)_{k_1},\cdots,x(n)_{k_n})) =  [x(1)_{{k_1}},x(1)_{{k_1}+1}[\times,$$
$$		 \cdots\times [x(n)_{{k_n}},x(n)_{{k_n}+1}[,$$
\begin{equation}
		 \forall k_1\in\mathbb{N}\cup[1,n_1[,\cdots,
		 \forall k_n\in\mathbb{N}\cup[1,n_n[ 
\end{equation}
We define $\mathcal{D}_{(x(1),\cdots,x(n))}$, the positive measure of data clouds associated with the set of all bins with the partitions $x(j)$. For a set of points discovered $H$ from $\mathcal{H}$, we have: 
$$\mathcal{D}_{(x(1),\cdots,x(n))}(H) = $$
\begin{equation}
 = \sum_{(k_1,\cdots,k_n)}{\textbf{1}_{\{\mbox{card} (bin(x(1)_{k_1},\cdots,x(n)_{k_n})\cap H)\gt 0\}}}
\end{equation}
For each dimension $j$, we choose the length of bin $l_{j}$ such that $\frac{|x(j)_{{k_j}}-x(j)_{{k_j}+1}|}{5} = l_{j}, \forall 1\leq j\leq n,\forall k_j$. $\mathcal{D}_{(x(1),\cdots,x(n))}$ is a well-defined positive measure for the associated set of non-empty bins of a data cloud $H$.

\begin{figure}[ht!]
	\centering
	\begin{minipage}{0.47\linewidth}
	\includegraphics[width=1.0\linewidth]{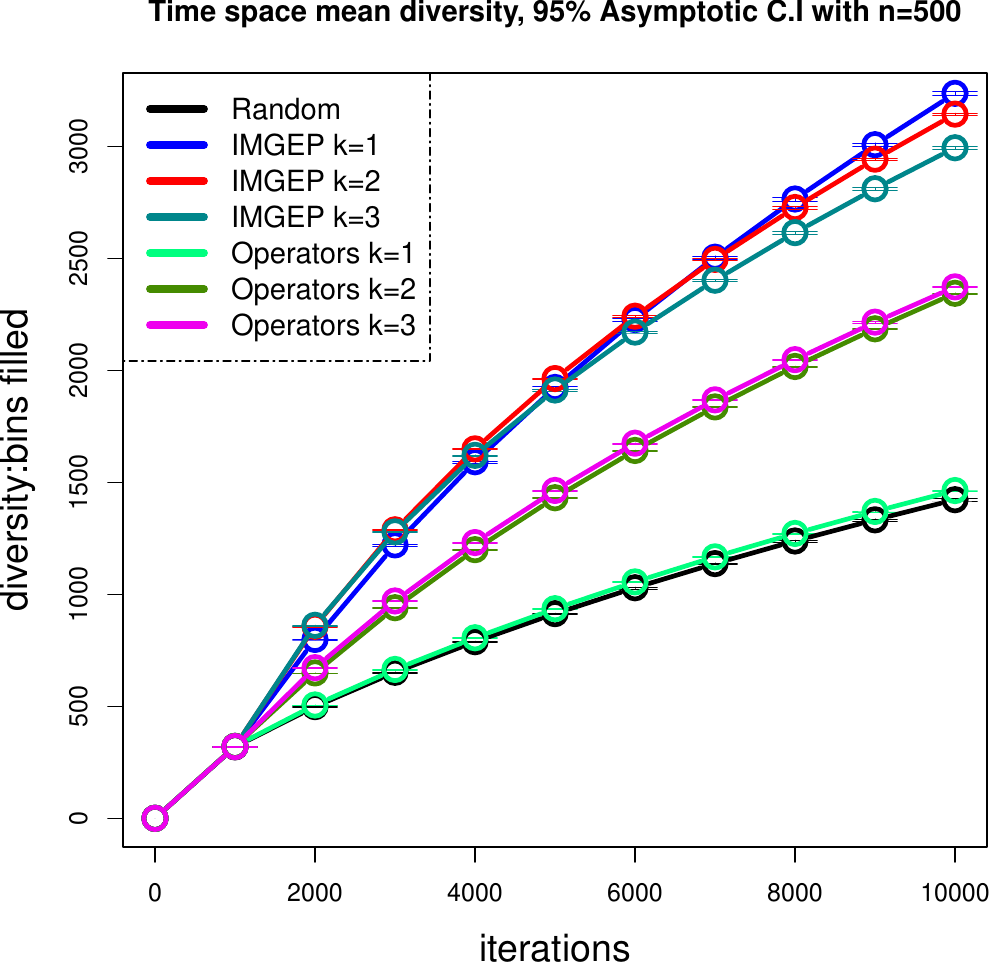}
	\end{minipage}
	\begin{minipage}{0.47\linewidth}
	\includegraphics[width=1.0\linewidth]{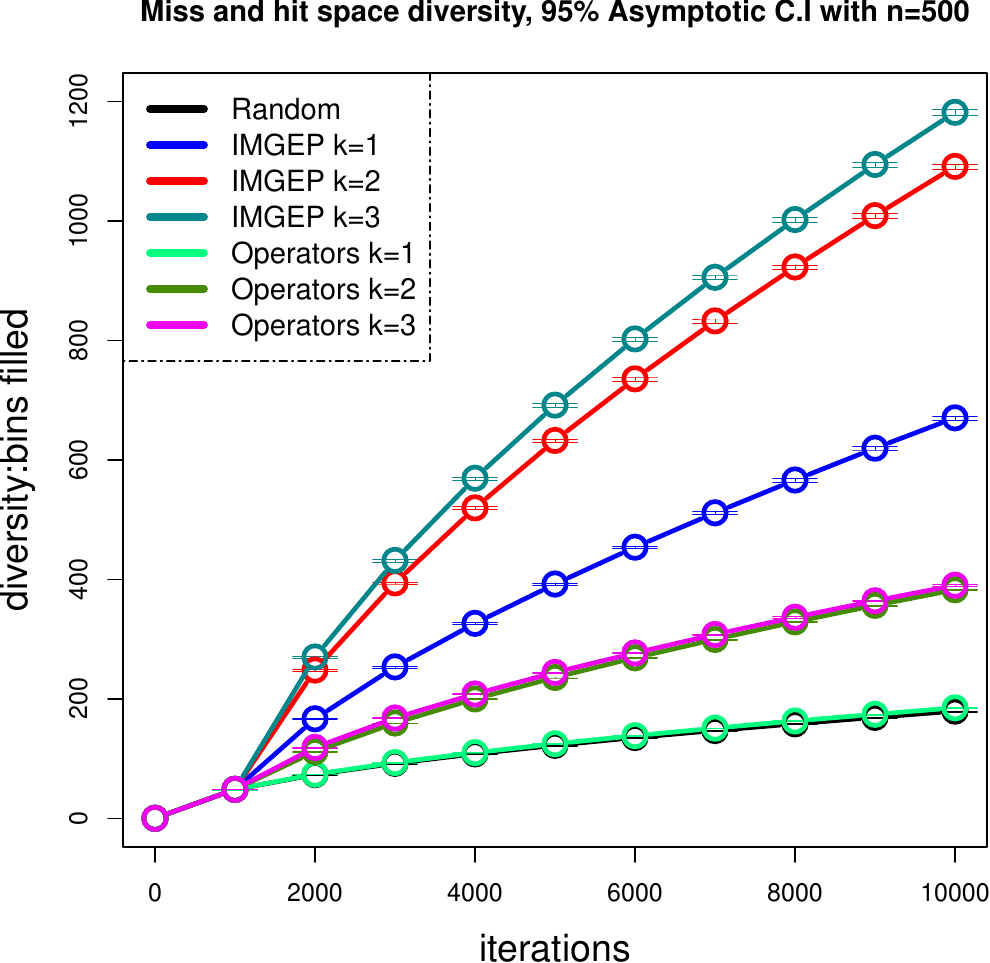}
	\end{minipage}
	\includegraphics[width=.5\linewidth]{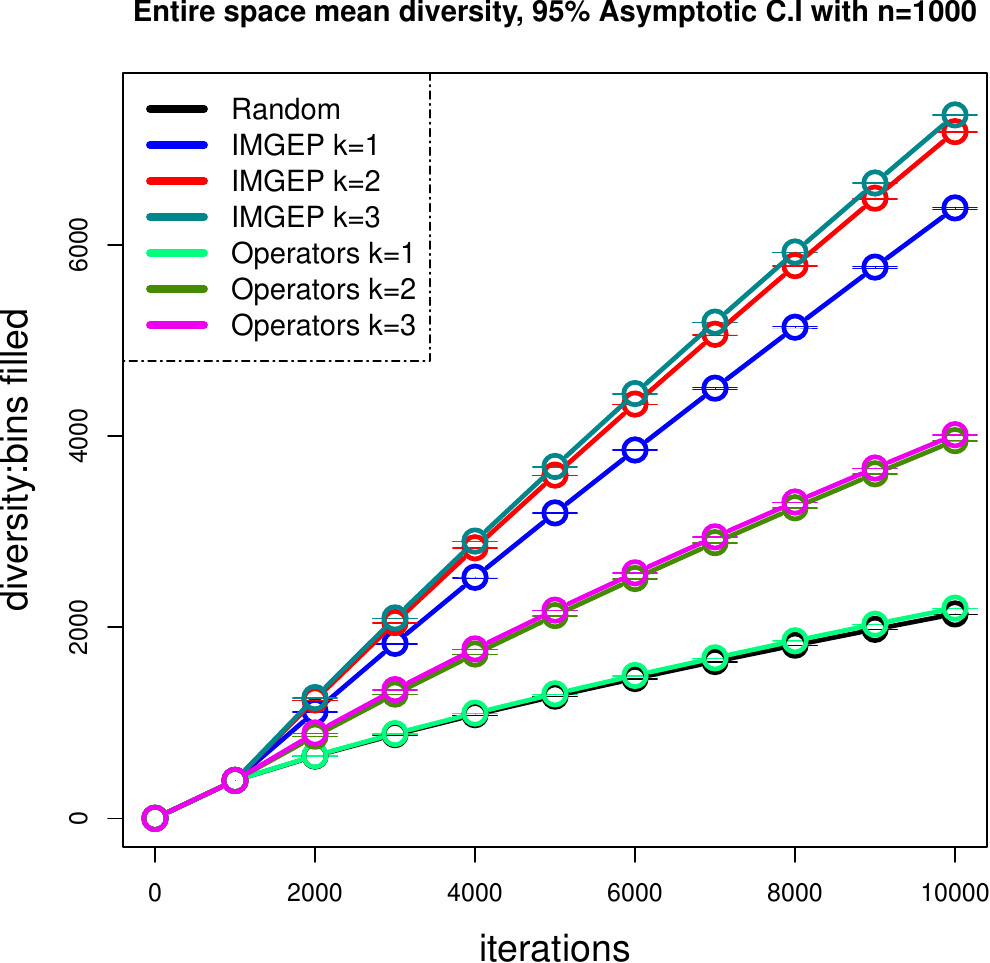}
	\caption{On the two left plots, we visualize the diversity on the time behavior space and on hit/miss space, that is, how the number of bins filled in a multidimensional histogram increases during exploration. On the right plot, we visualize the diversity of the entire dataset, the dimension of the considered histogram is the number of features (59). Results of $N=10^{4}$ iterations with a warming set of $N_{init} = 1000$ iterations of random program synthesis }

	\label{seperatediv}
\end{figure}
For each value of $k$, which is the number of neighbors in the optimization policy achievement model, we show that IMGEP leads to a higher diversity than the operator combination exploration.
Indeed, IMGEP finds many more distinct values of miss ratios across the distinct locations—pairs (bank, row)—in the main memory; see figure \ref{missratiosddr}. Note that the discoveries of "miss" occurrences probably help to induce more interference. Indeed, whenever requested data is missing in a DDR buffer, it has to be recharged, and this involves an additional delay. Moreover, figure \ref{seperatediv} shows that the IMGEP diversity measurement increases faster on the entire behavior space than that of the operator combination exploration, and simultaneously on the distinct L2 $\&$ DDR miss/hit and execution time goal subspaces. 

\begin{figure}[H]
	\centering
	\begin{minipage}{1.0\linewidth}
		\centering
		\begin{minipage}{.6\linewidth}
			\centering
			\includegraphics[width=1.0\textwidth]{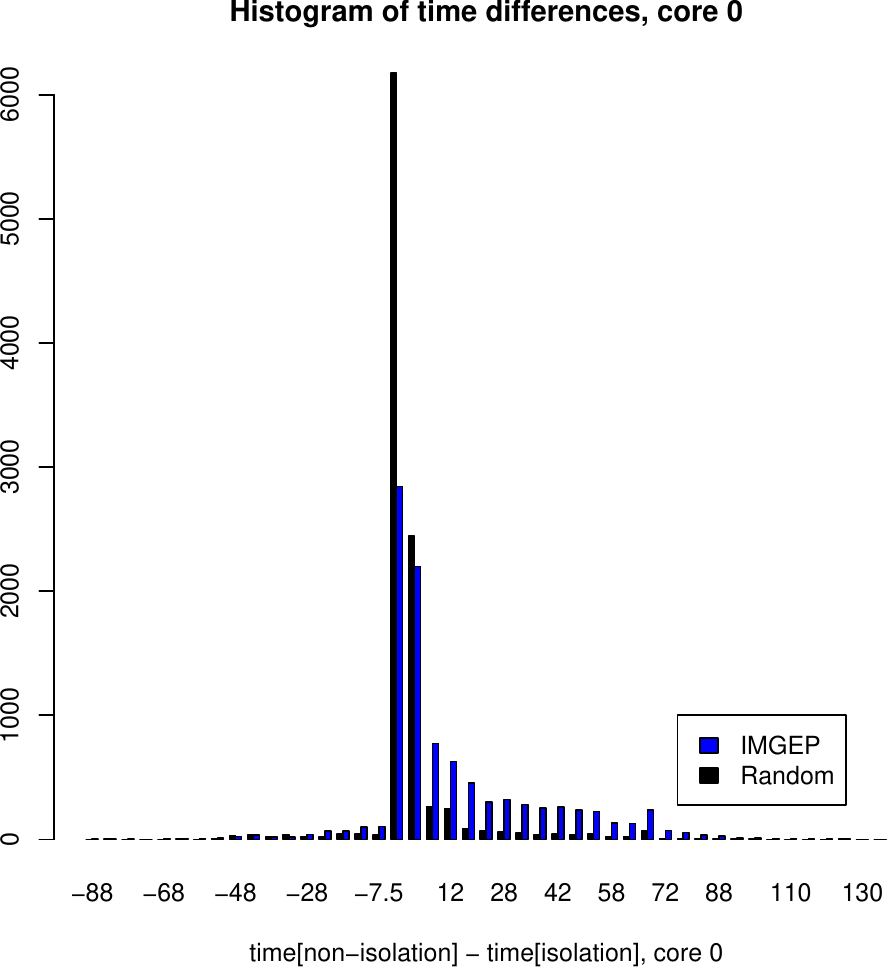}
		\end{minipage}
		\begin{minipage}{.6\linewidth}
			\centering
			\includegraphics[width=1.0\textwidth]{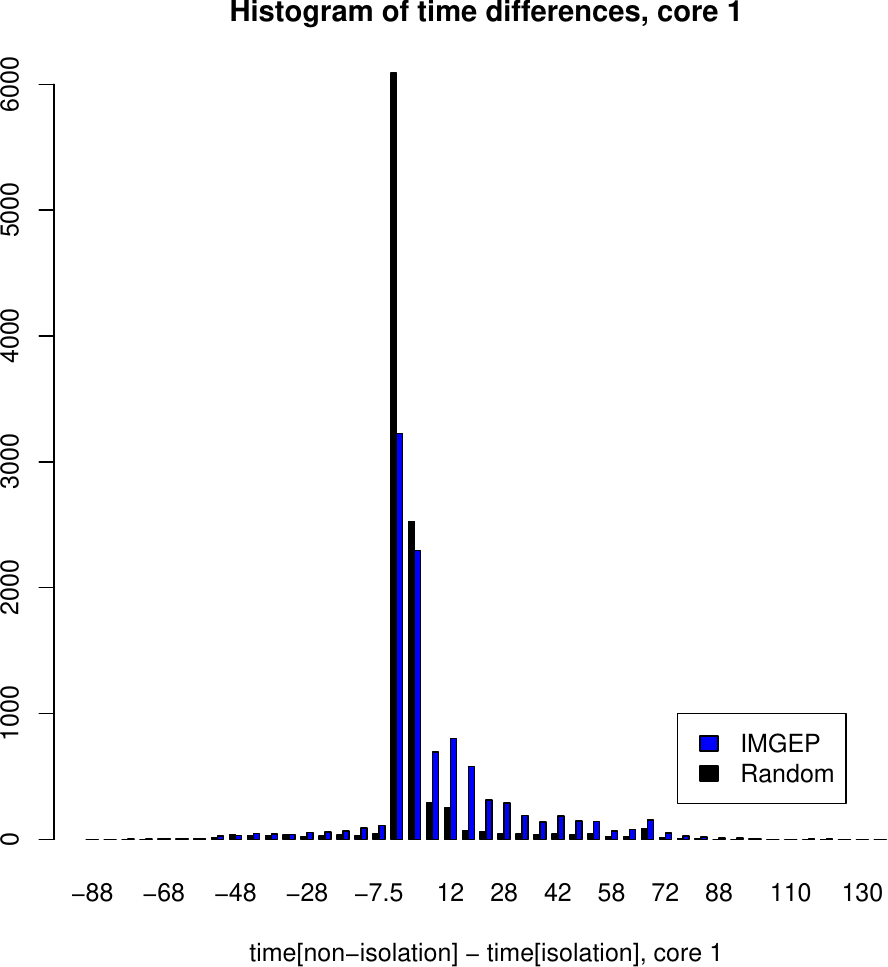}
		\end{minipage}
	\end{minipage}
	\caption{Diversity for execution time behavior space. Results of $N=10^{4}$ IMGEP iterations with $k=1$ and a warming set of $N_{init} = 1000$ iterations of random program synthesis.}
	\label{timediv}
\end{figure}

	Generating programs with IMGEP allows one to observe much more interference than generating programs randomly. Interference occurs whenever the execution time in isolation differs from the execution time in non-isolation. To some extent, we consider that a pair of programs produces interference if $|\mbox{time[non-isolation]} - \mbox{time[isolation]}|\gt0$. In the example run presented in figure \ref{timediv}, the bars with axis $0$ on histograms for values of |time[non-isolation] - time[isolation]| demonstrate that exploring the simulator with randomly generated programs produces interference about $4$ times out of $10$, whereas it is about $7$ times out of $10$ for IMGEP.

\begin{figure}[ht!]
	\centering
	\begin{minipage}{.45\linewidth}
		\centering
		\includegraphics[width=1.0\textwidth]{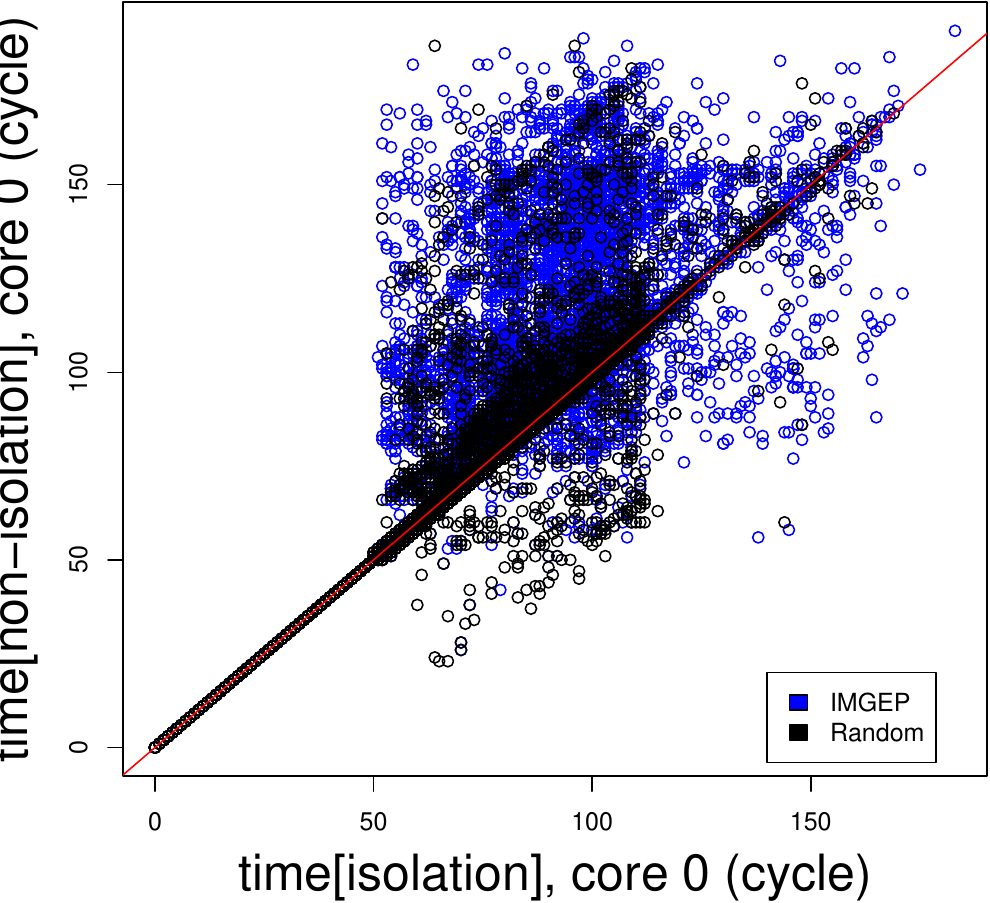}
	\end{minipage}
	\begin{minipage}{.45\linewidth}
		\centering
		\includegraphics[width=1.0\textwidth]{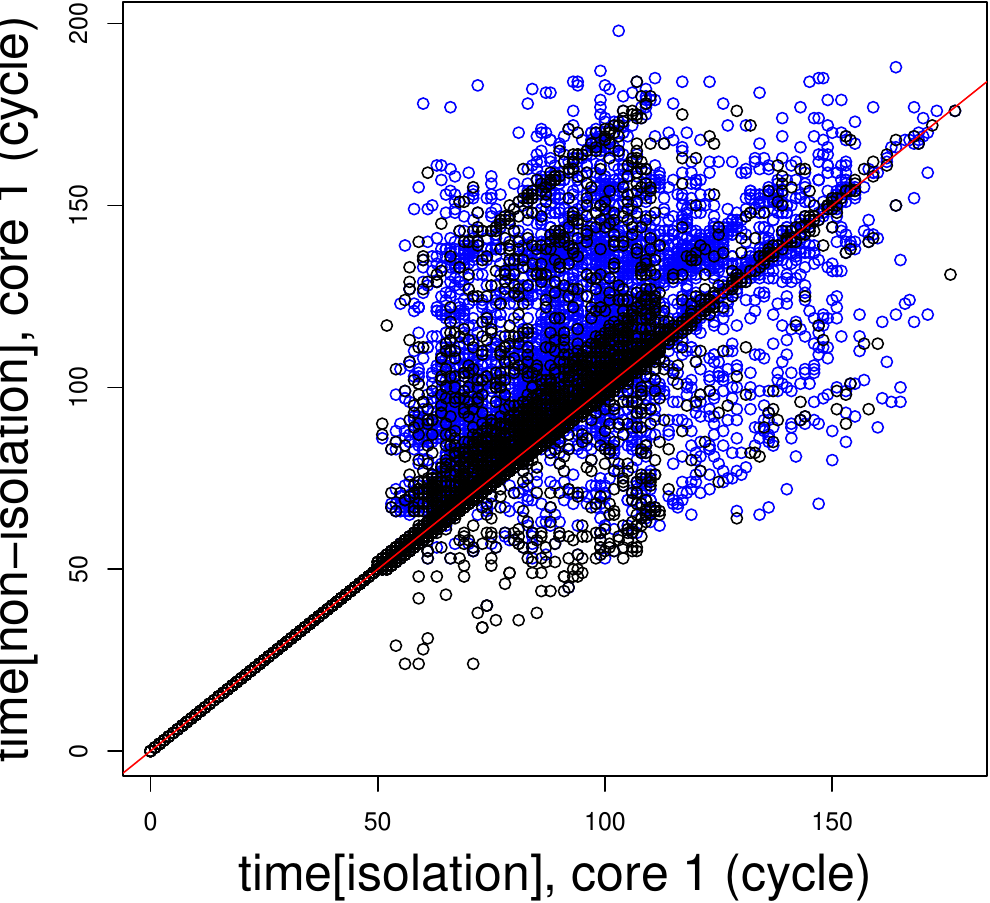}
	\end{minipage}
	\caption{Diversity for execution time behavior space. Results of $N=10^{4}$ IMGEP iterations with $k=1$ and a warming set of $N_{init} = 1000$ iterations of random program synthesis.}
	\label{timedivscatter}
\end{figure}

Information on the execution time is represented for each program in Figure \ref{timedivscatter}. More precisely, we attribute to each pair of programs executed simultaneously on core 0 and core 1 two respective points on the distinct scatter plots to represent its couple (time[isolation], time[non-isolation]). This representation allows us to spot interference, that is, whenever points are not located on the red identity line. Points above the red line represent programs that are slowed down by the parallel execution on the other core, whereas points below the line represent programs which are accelerated, which means the execution time is faster when the program is executed together with another program. This is due to the absence of a coherency management mechanism. For instance, the following pair of programs results in the acceleration of both programs. One can show DDR bank and line numbers instead of addresses. Instructions of core $1$ on cycles $36, 40$, and $41$ are processed earlier when the program on core 0 is executed because of its instructions on cycles $20$ and $30$ which load the lines $\{\mbox{bank } 1, \mbox{row } 1\}$ and $\{\mbox{bank } 0, \mbox{row } 1\}$ into their buffer. Hence, sequences PRE=$\gt$ACT are avoided as the data is already in the buffer.

\begin{lstlisting}[caption={Example of programs inducing mutual acceleration.},basicstyle=\ttfamily\scriptsize]
bk_rw_core0 =  {20:{'bk':1,'rw':1}, 27:{'bk':3, 'rw':0},
30:{'bk':0, 'rw':1},31: {'bk': 2, 'rw': 0},
36: {'bk': 1, 'rw': 0},43: {'bk': 3, 'rw': 0},
46: {'bk': 0, 'rw': 0}}

bk_rw_core1 = {6:{'bk':2, 'rw':1}, 8:{'bk':2, 'rw':1},
11:{'bk':3, 'rw':1},18:{'bk':1, 'rw':2},
27:{'bk':0, 'rw':2},36:{'bk':1, 'rw':1},
40:{'bk':1, 'rw':1}, 42:{'bk':0, 'rw':1},
45:{'bk':3, 'rw':1}, 47:{'bk':0, 'rw':2}}

core 1 mutuality 78 cycles, isolation: 162 cycles
core 0 mutuality 108 cycles,isolation: 97 cycles
\end{lstlisting}

\begin{figure}[ht!]
\centering
\includegraphics[width=.7\linewidth]{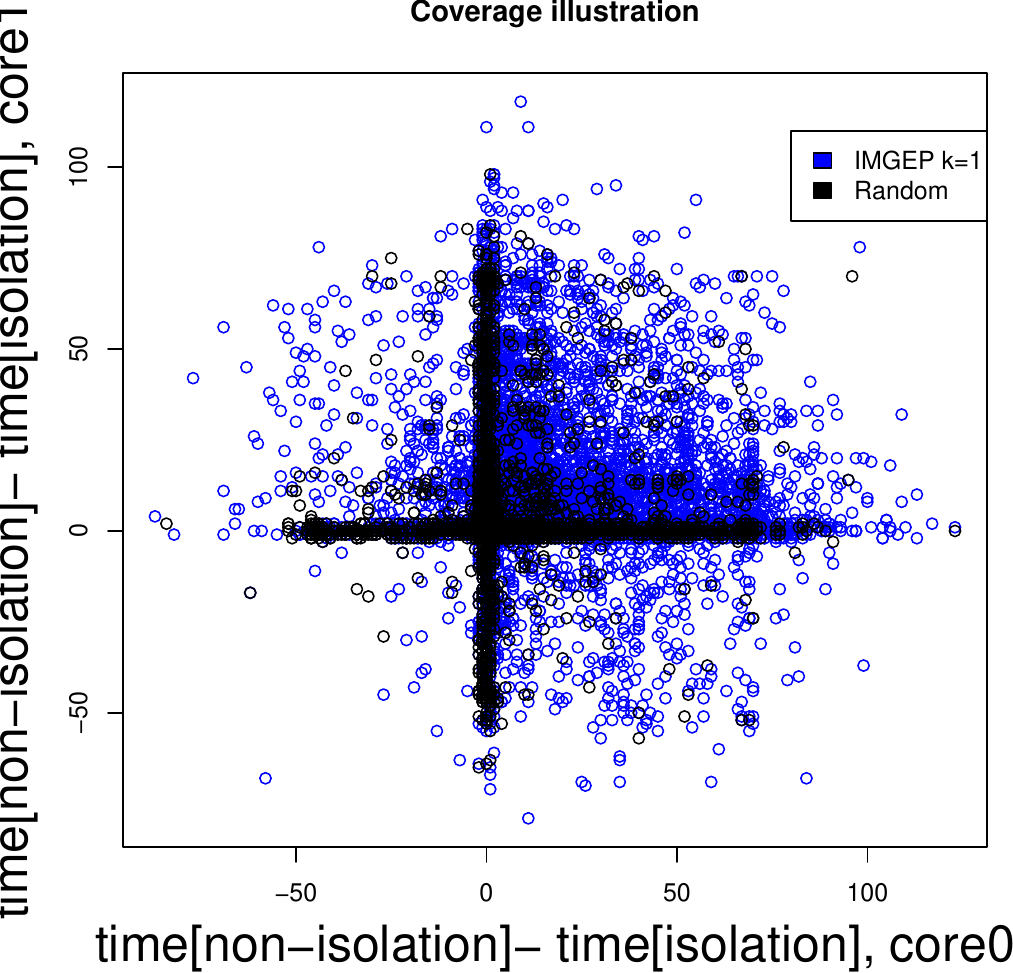}
\caption{Spanning of IMGEP on time behavior space. Interference occurs on core 0 or 1 when the abscissa or ordinate is different from zero. Results of $N=10^{4}$ IMGEP iterations with $k=1$ and a warming set of $N_{init} = 1000$ iterations of random program synthesis.}
\label{spanning}
\end{figure}

In Figure \ref{spanning}, we represent values of couples (time[isolation, core 0] - time[non-isolation, core 0], time[isolation, core 1] - time[non-isolation, core 1]). Any point with an abscissa or ordinate different from 0 corresponds to a pair of programs presenting interference on core 0 or core 1, respectively. The figure clearly shows that the IMGEP cloud spreads more, indicating that our algorithm finds many more interference cases.

\subsection{Analysis}
The results we presented demonstrate through diversity measurements that automated exploration algorithms can cover efficiently an output space of a complex system associated with a simulated hardware model.

The data from which the exploration is performed provides only aggregated information and is not internal.
While this implies that characterizing sources of interference with such a defined behavior space is limited, we spot an emerging structure in Figure 13, suggesting that pairs of programs with many requests to the same DDR locations induce a greater variety of interference delays.

Hence, the resulting pairs of programs causing interference can be further analyzed by humans, and our method may be completed to ensure that distinct points in the outcome space provide two distinct sources of interference.

\begin{figure}[ht!]
	\centering
	\begin{minipage}{.48\linewidth}
		\centering
		\includegraphics[width=\linewidth]{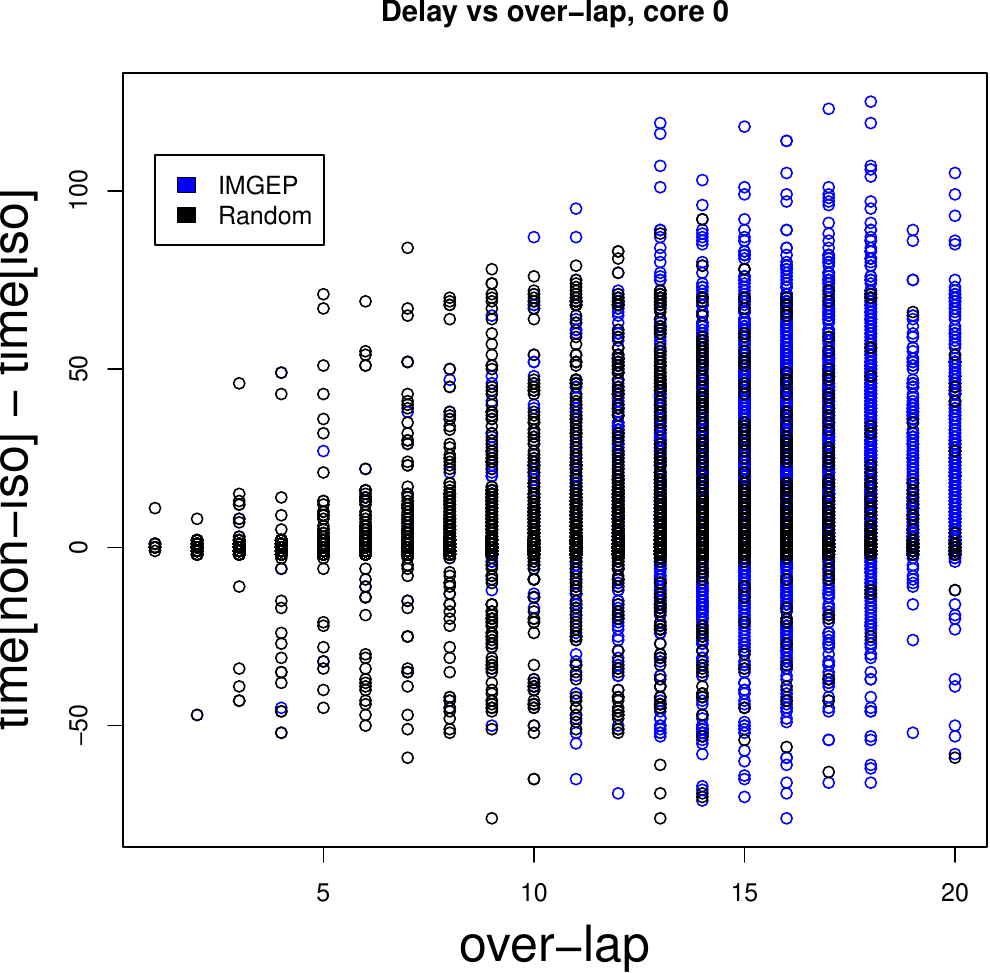}
	\end{minipage}
	\begin{minipage}{.48\linewidth}
		\centering
		\includegraphics[width=\linewidth]{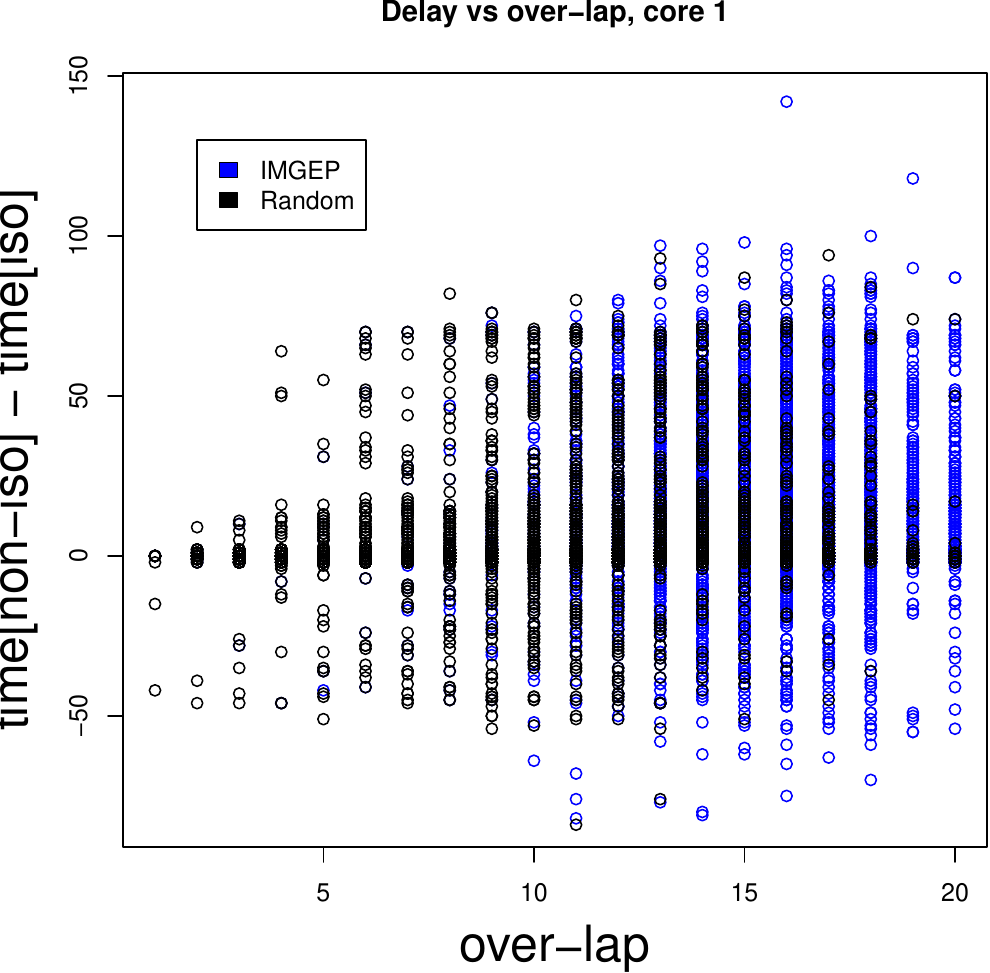}
	\end{minipage}
	\caption{(time[non-iso] - time[iso]) vs non-overlap. We call non-overlap the quantity $f(p_0,p_1) = \sum_{j} |(\mbox{nb access row j , core 0}) - (\mbox{nb access row j core 1})|$. This graph shows the number of access occurring in the same locations for both cores in the main memory is not enough to characterize interference.}
	\label{overlap}
\end{figure}
We believe that creating a space that fully consists of sources of interference will allow discoveries of non-trivial access patterns. In addition, one wants to discover access patterns that are unknown and difficult to produce by specialized engineers. Furthermore, the resulting behavior space would not necessarily be a Euclidean space, and could, for example, be a graph, a tree, or a semantic space.
\section{Conclusion}
This work deals with the problem of interacting with SoCs via automated discovery algorithms to help the discovery of singular cases. The method relies on simple machine learning models to structure an autonomous agent that fixes its own goals, and on the establishment of a behavior and a parameter space.

The results demonstrate that automated discovery algorithms can interact successfully with a simplified simulated model of dual-core architecture and encourage the idea of exploring a real SoC. Indeed, the behavior space, which consists of performance counters, is explored efficiently with IMGEP, as it leads to a higher diversity than exploring the simulator with programs that are randomly generated.

This approach encourages the idea of exploring a behavior space that is more suited for characterizing of mechanisms of interference in order to discover non-trivial sources of interference. In further work, we will aim to develop a strategy that specifically identifies distinct mechanisms that are responsible for interference, rather than exhibiting a large diversity of interferences regardless of the reasons they occur. For this, we will distinguish whether the observation of an interference relies on the activation of a new mechanism. Hence, the objective will be to maximize the number of distinct types of interference.

\enlargethispage{\baselineskip}

\begin{appendices}
	\section{Description of our dual-core model}

	\label{appendix:graphddr}

\begin{table}[ht!]
\centering
\resizebox{\linewidth}{!}{
\begin{tabular}{|l|l|l|l|}
\hline
\textbf{Component} & \textbf{Category} & \textbf{Description} & \multicolumn{1}{c|}{\textbf{Value}} \\ \hline
Core 0                & address range & address range for core 0 & 0-20  \\ \hline
Core 1                & address range & address range for core 1 & 21-40  \\ \hline
DDR                & bank mapping& Bank mapping for an address $a$& $a\%n_{banks}$  \\ \hline
DDR                & row mapping & Row mapping for an address $a$, each row covers 16 addresses & $a//16$ \\ \hline
	DDR                & tRCD (cycle)             & Delay induced by an ACT operation, that is, the activation of a closed line, by storing its content in the buffer & 15 \\ \hline
	DDR                & tRP (cycle)              & Delay induced by a PRE operation, that is, the deactivation of an open row for a given bank.                     & 15                                  \\ \hline
	DDR                & tCAS (cycle)             & Delay to access data after a burst for both RD and WR insttructions. & 15 \\ \hline
	DDR                & tFRC (cycle)             & Periodic time for cell refreshments. During this operation, every bank row buffer is precharged  & 30 \\ \hline
	DDR                & tWR (cycle)              & Minimum delay between a WR and a PRE operation   & 15 \\ \hline
	DDR                & tRTP (cycle)             & Minimum delay between an RD and a PRE operation   & 8  \\ \hline
	DDR                & tCCD (cycle)             & Minimum delay between an RD and a WR operation    & 4   \\ \hline
	DDR                & tRRD (cycle)             & Maximum delay between two ACT operations  & 4\\ \hline
	DDR                & $n_{\mbox{rows}}$             & Number of rows per bank & 3\\ \hline
	DDR                & $n_{\mbox{banks}}$             & Number of banks in the DDR& 4\\ \hline
L1,L2,L3 cache           & size              & Storage capacity of a cache (bytes) & 32,512,512\\ \hline
L1,L2,L3 cache           & line size         & Storage capacity of a line (bytes) & 4,4,4 \\ \hline
L1,L2,L3 cache           & associativity     & Number of different cache lines each data block can be mapped to. & 2,16,16  \\ \hline
\end{tabular}
}
\caption{Selection of the simulator parameters for our experiments}
\label{tab:parameters_simu}
\end{table}
\subsection{Complementary information on the DDR and its controller}
Our simplified DDR2 model implements the following operations and some are described in \ref{tab:parameters_simu}
	\begin{itemize}
		\item WR: Write
		\item RD: Read
		\item ACT: Activation of a closed row.
		\item PRE: Deactivation of an open row.
		\item REF: Refreshment 	of memory cells.
	\end{itemize}
		\label{appendix:ddrcontroller}
The functionning of both the DDR and the memory controller are coupled.
At each cycle:
	\begin{itemize}
		\item The controller determines requests that are achieved, that is, if the current date is superior than the scheduled completion date. In that case, for an RD request, the controller calls a "call back" function which signals to the core that this access is achieved. This callback is used by the core to allow the execution of a new instruction. No callback functionality is not implemented for RD requests.
		\item The controller treats requests that are located in its entry queue. For each request, it determines whether it can be treated by the DDR according to its state (handled by the DDR state machine) while ensuring minimal delay contraints.  It then determines the "best" request to treat depending on a established ranking to prioritize request inducing "row hits", WR over RD requests and arrival order (FIFO policy). THhe controller finally sends the "best" request to the DDR. Memory accesses are done with burst lenght of $8$ bytes.
	\end{itemize}
	The model is inspired from mechanisms described in \cite{gonzalez:tel-05025151}.

\subsection{Interconnect model}
	The interconnect model can be described as follows:
	\begin{enumerate}
		\item It consists in maintainig a queue (FIFO) of memory requests and in executing request in the arrial order and after a delay corresponding to minimale access latency. A random relay of 2 cycles in added.
		\item At each clock cycle, some number of requests in priority is unpiled from the queue among the ones that are executable during the current cycle. The requests are the, sent to the memory. The number of unpiled requests at each cycle is determined by the bus bandwidth and each the counting of the numberf of treated requests.
		\item The requests that could not be treated are moved in the queue of requests that are to be treated in next cycles.
		\item The interconnect clock cycle frequency and that of the cores are the same. One could introduce distinct clocks by calling the functions "clock" in our codes at  distinct cycles.
	\end{enumerate}
\section{Intrinsically Motivated Goal Exploration Process}
\label{appendix:IMGEP}
Parameters chosen for the IMGEP exploration are described on table \ref{tab:IMGEPparam}.
\begin{table}[ht!]
\centering
\resizebox{\linewidth}{!}{
\begin{tabular}{|l|l|l|}
\hline
\textbf{IMGEP parameters} & \textbf{Description} & \multicolumn{1}{c|}{\textbf{Value}} \\ \hline
$N_{init}$& Number of iterations for the warming set& 1000\\ \hline
$N - N_{init}$& Total number of IMGEP iterations & 9000\\ \hline
$k$& Number of neighbors in the $kNN$ model for the goal achievement strategy $\Pi$. & 1,2,3\\ \hline
Number of mutations & Number of mutation for each program in the model $\Pi$& 5\\ \hline
max sending cycle   & Maximum cycle for sending the instruction (cycle)& 60\\ \hline
Range instructions  & Minimal and maximal number of instructions& 1-10\\ \hline
core 0& Range of addresses & 0-20 \\ \hline
core 1& Range of addresses & 21-40 \\ \hline
\end{tabular}
}
\caption{The selection of IMGEP parameters for our experiments.}
\label{tab:IMGEPparam}
\end{table}
\label{appendix:mixing}
We describe below the algorithm designed to mix sequences of simplified assembly algorithms. As a reminder, the algorithms are dictionaries of form $\{cycle:(type,adress)\}$.
\begin{algorithm}[ht!]
\caption{Random Mixing of Instruction Sequences}\label{mix_sequences}
\footnotesize
\begin{algorithmic}%[1]
\Require A list of instruction programs $S$, number of parts $P$, random seed $s$, maximum cycle $C_{\max}$
\Ensure A mixed instruction program $M$
\State Initialize random generator with seed $s$
\State $Chunks \gets \emptyset$
\ForAll{program $p \in S$}
    \State Sort instructions of $p$ by increasing cycle
    \State Split $p$ into $P$ contiguous parts
    \ForAll{part $q$ of $p$}
        \State Append $q$ to $Chunks$
    \EndFor
\EndFor
\State Randomly shuffle $Chunks$
\State Concatenate all elements of $Chunks$ into list $L$
\State Randomly select $|L|$ distinct cycle values from $[1, C_{\max}]$
\State Sort selected cycle values in increasing order
\For{$i \gets 1$ to $|L|$}
    \State Assign cycle $cycles[i]$ to instruction $L[i]$
\EndFor
\State Construct mixed program $M$ from assigned cycles and instructions
\State \Return $M$
\end{algorithmic}
\end{algorithm}

\label{appendix:mutation}
\begin{algorithm}[H]{
    \caption{Mutation operator}}\label{alg-mutation}
    \begin{algorithmic}%[1]
	    \scriptsize
\State // Mutate an instruction sequence by adding,     deleting, or modifying instructions.
\Require Instructions $\theta$, Number of mutations $N$, Maximal cycle $c$, range for adresses $a_{min},a_{max}$,maximal number of instructions $i_{max}$

\Ensure Initialize a copy $\theta^{'}$of the intructions $\theta$
\Ensure Initialize set of available cycles \textbf{freecycles}: \textbf{all cycles} - \textbf{used cycles}
\For{$i \leftarrow 1\colon N$}
	    \If{lenght$(\theta^{'})>1$}
	    	\State Select a mutation type : $t \sim \mathcal{U}([\mbox{add}, \mbox{delete}, \mbox{modify}])$
	\Else
	    	\State Select a mutation type : $t \sim \mathcal{U}([\mbox{add}, \mbox{modify}])$
	\EndIf
	    \If{mutation type is add and \textbf{freecycles} non empty}
		\State \textbf{newcycle} $\sim \mathcal{U}$(\textbf{freecycles})
		%\State \textbf{instruction type} $\sim \mathcal{U}([\mbox{read},\mbox{write}])$
		%\State address $a = \mathcal{U}([a_{min},a_{max}])$
		\State $\theta^{'}(\textbf{newcycle}) =$  \{\textbf{new instruction type},$a$\}
		\State remove \textbf{freecycles}(\textbf{newcycle})
	\ElsIf{mutation type is delete and lenght$(\theta^{'})>1$}
		%\State \textbf{cycle to delete} $\sim \mathcal{U}(\theta^{'})$
		\State Delete $\theta^{'}(\textbf{\mbox{cycle to delete}})$
		\State set free \textbf{freecycles}(\textbf{cycle to delete})
	    \ElsIf{Mutation type is modify and lenght$(\theta^{'})>1$}
	    %\State cycle to modify $c\sim \mathcal{U}(\theta^{'})$ 
	    \State Pick an instruction cycle to modify $t_{old},a_{old}= \theta^{'}$(c)
	    \State modify choice $m \sim\mathcal{U}([\mbox{type},\mbox{adress},\mbox{both}])$
	    \If{$m =$ type}
	    	\State $t_{new} =$ {write }\textbf{if} $t_{old}==\mbox{read }$\textbf{else} read
		\State $\theta{'}(c) = \{\mbox{type}\colon t_{new},\mbox{address}\colon a_{old}\}$
	    \ElsIf{$m$ address}
		%\State new address $a_{new}\sim\mathcal{U}(\{a_{min},\cdots,a_{max}\})$
		\State $\theta{'}(c) = \{\mbox{type}\colon t_{old},\mbox{address}\colon \mathcal{U}(\{a_{min},\cdots,a_{max}\})\}$
		\Else
		\State //Change both type and address
	    	\State $t_{new} =$ {write }\textbf{if} $t_{old}==\mbox{read }$\textbf{else} read
		%\State new address $a_{new}\sim\mathcal{U}(\{a_{min},\cdots,a_{max}\})$
		\State $\theta{'}(c) = \{\mbox{type}\colon t_{new},\mbox{address}\colon \mathcal{U}(\{a_{min},\cdots,a_{max}\})\}$
	    \EndIf
	\EndIf
	    \If{lenght$(\theta^{'})>i_{max}$}
	\State Uniformally sample set of distinct lenght$(\theta^{'})-i_{max}$ instructions to delete
	\State Delete the instructions
	\EndIf
\EndFor
    \end{algorithmic}
\end{algorithm}

\subsection{Diversity Evaluation}
Another diversity measure, but ill-defined, is the sum of the squared euclidean distances between all pairs: $\mathcal{D}(\mathcal{H})=\sum_{i}\sum_{j\leq i}{||o_i - o_j||}^{2}$. This measure is not a well defined positive measure as $\mathcal{D}(H_1\cup H_2)\neq\mathcal{D}(H_1)+\mathcal{D}(H_2)$, when $H_1\cap H_2=\emptyset $ in general. However, it provides a good intuition of the diversity, as a small perimeter cloud with small pairwise distances leads to a small diversity result whereas a large cloud perimeter involves larger pairwise distances and thus a larger diversity.
\end{appendices}

\printbibliography
\end{document}